\documentclass[journal]{IEEEtran}

\usepackage[noadjust]{cite}
\usepackage{amsmath}
\usepackage{amsfonts}
\usepackage{amssymb}
\usepackage{amsthm}
\usepackage{array}
\usepackage{graphicx}
\usepackage[T1]{fontenc}
\usepackage{subfig}
\newtheorem{theorem}{Theorem}
\usepackage{lineno, blindtext}

\usepackage{graphicx}
\usepackage{cleveref}
\usepackage[]{units}
\usepackage[ruled,lined,boxed,commentsnumbered]{algorithm2e}
\newcommand{\argmin}{\arg\!\min}
\usepackage{color}

\usepackage{slashbox,pict2e}

\theoremstyle{definition}

\usepackage[usenames,dvipsnames]{pstricks}
\usepackage{epsfig}
\usepackage{pst-grad} 
\usepackage{pst-plot} 
\usepackage[space]{grffile} 
\usepackage{etoolbox} 
\makeatletter 
\patchcmd\Gread@eps{\@inputcheck#1 }{\@inputcheck"#1"\relax}{}{}
\makeatother

\begin{document}
\title{An Integrated Decision and Control Theoretic Solution to Multi-Agent Co-Operative Search Problems}
\author{Titas Bera, Rajarshi Bardhan, Sundaram Suresh,
\thanks{*This work was supported byt ST-Engineering under project CRP3-P2P}
\thanks{Titas Bera is with the School
of Electrical and Electronic Engineering, Nanyang Technological University, Singapore,
e-mail: btitas@ntu.edu.sg}
\thanks{Rajarshi Bardhan is with the School
of Electrical and Electronic Engineering, Nanyang Technological University, Singapore,
e-mail: rbardhan@ntu.edu.sg }
\thanks{Sundaram Suresh is with the School
of Computer Science and Engineering, Nanyang Technological University, Singapore,
e-mail: Ssundaram@ntu.edu.sg }}
\maketitle
\begin{abstract}
This paper considers the problem of autonomous multi-agent cooperative target search in an unknown environment using a decentralised framework under a no-communication scenario. The targets are considered as static targets and the agents are considered to be homogeneous. The no-communication scenario translates as the agents do not exchange either the information about the environment or their actions among themselves. We propose an integrated decision and control theoretic solution for a search problem which generates feasible agent trajectories. In particular, a perception based algorithm is proposed which allows an agent to estimate the probable strategies of other agents' and to choose a decision based on such estimation. The algorithm shows robustness with respect to the estimation accuracy to a certain degree. The performance of the algorithm is compared with random strategies and numerical simulation shows considerable advantages.
\end{abstract}

\IEEEpeerreviewmaketitle
\section{Introduction}
It is already an established method to deploy robots to accomplish certain tasks in various application scenario such as 
reconnaissance and surveillance operations, damage assessments, space exploration, and scientific data gathering etc,. In general, applications which considered to be costly and dangerous for human operatives naturally admit robotic installations. In recent years, an interest is grown to deploy a team of agents or robots or UAVs, instead of a single robot, to achieve more operational capabilities. A team of robots, under a distributed decision making framework, also offer more robustness from a reliability point of view. Such distributed framework often classified as distributed multi-agent systems, where each single agent makes a decision locally based on neighbourhood information, while trying to achieve the overall global mission objective together with the other agents in the system. For example, a group of UAVs, having a limited range of sensing capabilities, are deployed in a previously unknown area, trying to minimize a global environmental uncertainty measure while doing a local area search, in an efficient co-operative manner.

The general problem of such target search and unknown area surveillance using distributed or decentralised co-operative multi-agents is well known. Early works such as \cite{gage1992command}, and \cite{gage1994randomized} consider such problem. In a typical instance of a multi-agent search problem, the objective is to simultaneous trajectory planning of multiple UAVs in order to search an environment for possible targets.  The environment model is unknown to each agent and it can be assumed that each agent only possesses an approximate initial uncertainty map measure of the environment. In the general setting, during the course of search operation, the agents can exchange communications, as well as merge informations in either in a centralized \cite{sujit2004search} or in a decentralized \cite{Passino2002} manner. Decision algorithms, centralised or decentralised, then use those informations to generate action for each agent such that, the global objective of identifying possible targets and minimization of the environmental uncertainties, is achieved. This can be achieved, for example, by exploiting the relationship between local and global agent objective cost functional \cite{terelius2011decentralized}. However this requires exchange of informations between the agents, that is, there must exist a well defined communication protocol and structure.


In this paper we considered the problem of target search and area surveillance by UAVs in a more complex setting, that is, under no-communication scenario. Such a scenario is a very practical one as it often arises when there is a limitation in the bandwidth in the communication protocol, or, the area should be surveyed in complete radio silence. Consideration of such a no-communication scenario also implies that the UAVs, equipped with standard instrumentation gauge and range sensors, are less burdened with communication hardware and consequently represents a low cost solution and more endurance. However, with no-communication scenario the problem trajectory planning of UAVs in co-operative search mode, becomes extremely challenging as the incremental environmental informations and control actions of the UAVs are not getting exchanged between the agents.

As without any exchange of informations, simultaneous co-operative trajectory planning is an impossible task, in this paper, we propose an alternative decentralized perception based co-operative search framework, which is based on an agent's perception (and not the true state) about the environment as well as on other agents' actions. To substantiate the rationality behind such approach, we refer to \cite{ho1972team}, where it is shown that an agent's decision on its action, under no-communication scenario, only depends on the then perceived environmental conditions. Therefore, in such cases, the agent's decision does not depend on the other agents' action, hence each agent chooses a greedy approach and as a result of which any expected synergy is lost in the process. To circumvent this problem, therefore, we allow not only the agents to estimate the state of the environment but also to guess or perceive each others' action and to include those in minimizing a individual local objective function. 

Interestingly, although a game theoretic argument can be given to show that such perception based approach can lead to a 'safe' strategy in an average sense and show a co-operative behaviour for target search, it is evident that an exhaustive enumeration of other nearby UAVs' action, by a single UAV, is essentially having an exponential complexity with the number of UAVs \cite{yang2004decentralized} and not scale free. Therefore, to circumvent such curse of dimensionality, we use randomized techniques for trajectory generation . In particular each UAV estimates a most probable action of the other UAVs and consequently acts upon it. It is clear that this by no means gives a guarantee on future expected behaviour of the neighbouring UAVs, however, we show, by Monte-Carlo simulation, that inaccuracies in the estimation, to some degree, do not have a drastic effect to the performance of the overall search. That is, the proposed method enjoys a certain inherent robustness. 
Note that, a reduction in the enumeration of UAV actions can also be achieved by creating a very course representation of the environment, such as a discretizing of the search area into rectangular grids, and limiting the number of possible actions that a UAV can execute. This discretization, although, severely limits the effectiveness of the co-operative search, a clear benefit of the above process is imminent in cases where the decision and control execution layer enjoys a clear separation principle, that is, there exists a large time scale separation for decision and control loop to execute. Existing solutions \cite{flint2002cooperative}, \cite{yang2004decentralized} to the general problem of target search and area surveillance considers such decoupled algorithmic approach. These algorithms consist of an upper level decision theoretic planning layer and an inner level control theoretic execution layer. The decision layer generates a trajectory ignoring an agent's dynamics. Such trajectory planning calculation is therefore only based on a pure kinematic representation of the overall system. Subsequently an appropriate controller is designed which ensures that the agent follows the trajectory, generated in the first phase, while satisfying agent's own dynamics. One can see that such decoupled approach has several disadvantages. First, the path generated at the planning stage may not be feasible or attainable due to the vehicle dynamics. Secondly, in the presence of obstacles, to generate a collision free trajectory solution, an explicit use of system dynamics is necessary. For example, a configuration, consisting of position and orientation of an agent, may not be in a collision condition with any of the environmental obstacles, but the magnitude and direction of velocity may be such that a collision is imminent. In other words, the agent may be in a collision course. This is quite common for systems with significant limits on available controls. These implies an integrated decision and control theoretic framework is necessary for multi-agent target search and surveillance problem. In other words, the integrated approach should take the kinematic representation of the over all system as an input and consequently generate control commands as output. Note that such approach alleviates necessary discretization of the environment and consequently maintains the continuum natures of the search environment.

Overall, we proposes a randomized algorithm based integrated decision and control (IDC) theoretic approach to multi-UAV co-operative search problem under no-communication constraint. The corresponding IDC-algorithm does not require any grid data structure of the underlying search area and can generate UAV trajectories that are feasible and attainable. The resulting co-operative search enjoys a robustness to some degree with respect to the perception inaccuracy of the individual UAVs. We also show that the randomized IDC-algorithm is probabilistically complete.

In the following, in section \ref{Related Work}, we give a brief overview of the existing literature is given. Section \ref{Cooperative Search Framework} outlines the problem of decoupled decision and planning and discuss the IDC framework. Section \ref{Problem Formulation} contains details of the target search and surveillance problem formulations. Section \ref{Integrated Planning} outlines the randomized Integrated decision and control algorithm.
Section \ref{Simulation} shows example simulation of the proposed method and discuss the various results. Section \ref{Analysis} shows the probabilistic 
completeness analysis of the proposed search method. Section \ref{Conclusion} summarize the findings and conclude.

\section{Related Work}\label{Related Work}
Multi-Agent co-operative decision making consists of decision making and information merging. Both these can be done in a centralised or in a de-centralised mechanism. In this section, we will briefly outline relevant works which uses both of these mechanisms. For a more comprehensive literature survey, see \cite{cao2013overview}.

A cooperative search framework for MAVs, first of its kind, to reduce uncertainty  while increasing coverage efficiency of the search region was proposed in \cite{Passino2002}.  Most of the early stage literature in this field were based on centralized maps with information merging, but employ distributed decision making for movement \cite{flint2002cooperative}. A common feature among these methods is that a UAV calculates its own path using the search map and measurements/estimates of some state variables of interest related to other UAVs in the team. A variety of techniques like artificial potential fields \cite{yang2007multi}, machine-learning techniques \cite{yang2004decentralized}, group dispersion patterns \cite{york2012ground}, mixed-integer linear programming \cite{forsmo2013optimal}, and evolutionary algorithms \cite{berger2010co} have been applied in this domain to demonstrate their improved search efficiency  by reducing the overlap in look-ahead planned paths. Information-theoretic sensor management was used in \cite{kolba2011framework} to propose a  distributed decision making based on a centralized search map, taking false alarms and miss-detections in the sensor model into consideration. The proposed model suggests UAV to move on a grid cell to maximize the expected 
information gain by future sensor observations. Although useful in reducing the uncertainty about the search region,  
this method did not consider sensing and communications constraints arising in real world scenario. The works in  \cite{enns2002guidance}, \cite{yanmaz2010stationary} achieved promising results in finding the target locations and distributed coordination of decision making even without having a search map. However, \cite{enns2002guidance} neither consider communication or sensing radius limitations, whereas \cite{yanmaz2010stationary}  considers communication range limitations but no sensor range limitation.

The works in \cite{bourgault2003coordinated}, \cite{tisdale2009autonomous}, \cite{delle2010decentralised} proposed a concept of building a distributed search map by sharing sensor observations. This method requires each MAV to update its own search map and search action individually to locate the target efficiently. This shows how efficiently information about a target location can be maintained in a distributed manner within an MAV team. However, these methods do not consider communication limitations and false alarms in their sensor model. In \cite{sujit2011self}, the proposed technique in distributed information merging and decision-making strategy  involves exchange of multiple messages. Although this method assumes limitations in communication, it does not consider a realistic sensor model into account. Another work \cite{chung2008multi} proposed distributed information merging and decision-making 
where the UAVs share binary sensor observations to coordinate. Though this method considers a realistic sensor model with both types of errors, it does not include limitations in communications. The work in  \cite{hu2013multiagent} in this category of cooperative search focuses on consensus among UAVs to maintain similarty in the maps of each UAV with a finite number of observations, while considering the limitations in communication and sensing (both types of errors) operation.

While information merging can be either centralized or distribute, decision making can also be centralized \cite{sujit2004search}, \cite{lum2010search}, \cite{mirzaei2011cooperative} or distributed \cite{gil2008stable}. In the method proposed  in \cite{gil2008stable} the UAVs can merge information in terms of collecting observations from other team mates. A technique that uses centralized decision making without any information merging was introduced in \cite{riehl2011cooperative}. In this method, a centralized entity repeatedly assigns the UAVs to the subregions that need to be visited, without any merging of information. The methods discussed so far has a common attribute that they all try to reduce uncertainty within a given cell by increasing the number of observations in it. 

\section{Integrated Decision and Control Framework}\label{Cooperative Search Framework}
\begin{figure}
  \centering
   {\includegraphics[scale = 0.8, keepaspectratio = true]{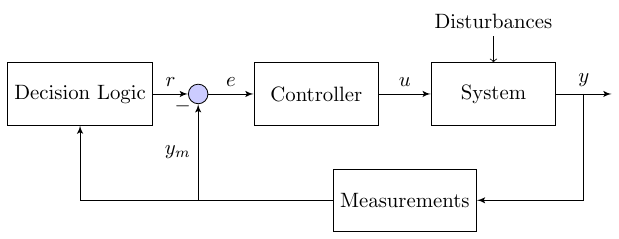}}
  \caption{General cooperative search framework}
  \label{Decoupled}
\end{figure}

In general the cooperative search framework consist of one outer decision layer and an inner control execution layer, as shown in Figure \ref{Decoupled}. The outer decision layer is mainly responsible to generate trajectories to be followed by each agent. In inner control layer, a trajectory following controller achieves the desired trajectory in spite of environmental disturbances. Often the implicit understanding is that in such decoupled scheme there exists 
a time scale difference between the path planning problem and trajectory following problem, so that these two problem can separately be considered. There also exists a representational difference between the two layers. For example, the outer loop path planning or trajectory generation loop considers a very simplistic representation of agents. Considering an UAV as an agent, the position and heading are considered to be slow moving dynamics when compared to pitch or roll stabilization of the UAV.

Therefore the decision layer of the agent operates on a minimalistic representations of self as well as other agents and only reacts to large scale environmental changes. The decision layer of each agent exchanges information about the environment and updates the cognitive map of the same. In this setting the cooperation implies maximization of a 
global benefit defined over the environment. The type of co-operation considered here is a passive co-operation in the sense that all agents are completely autonomous. Neither the agents tell other agent as what to do, nor there exists a hierarchy of structure or negotiations between the agents.

However, when considering this decoupled approach, the outer decision layer may demand a trajectory following control law which may not be feasible. The agents often have control or actuation limitations. Moreover, such trajectory following control generally becomes a two point boundary value problem which may not posses a solution because of control constraints. For example, in case of UAVs as agents, there may be a restriction on its turn rate and control surface deflections. Such constraints and non-feasibility affects the overall trajectory of the agent. In such cases, the Frechet distance between the desired trajectory and the actual trajectory can be very high. Moreover, the agent undergoes a motion, some part of which falls into restricted reachable sets or obstacles that the decision layer did not take into account. One solution could be to maintain a sufficient distance from the obstacles while planning for a trajectory, but this will only make the solution conservative.

\begin{figure}
  \centering
   {\includegraphics[scale = 0.8, keepaspectratio = true]{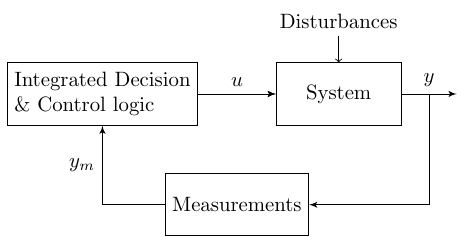}}
  \caption{Integrated Decision and Control framework}
  \label{Coupled}
\end{figure}

Therefore it is important to consider simultaneous decision planning and control of the agents exercising a co-operative search. This integrated approach must consider simultaneously the kinematics based representations of the overall scenario and dynamics of the individual system, that is, it must take into account the overall kino-dynamic problem, see Figure \ref{Coupled}. In this context, a general formulation of the problem of integrated multi-agent system decision theoretic control is presented. Note that, the integrated approach is a generalization of the multi-layered approach.


With this perspective in the following section we introduce the problem of multi agent search in detail.
\section{Problem Formulation}\label{Problem Formulation}
\subsection{A Single Agent}
Before describing the multi-agent framework, it is necessary to start with a single agent. An agent can be a single robotic manipulator in a production pipeline, an unmanned aerial vehicle in a search and reconnaissance mission, or an underwater vehicle. If the agent has $d$ degrees of freedom, then it can be represented as a point in a $d$-dimensional space, called the configuration space $\mathbb{C}$, which is locally equivalent to a $d-$dimensional Euclidean space $\mathbb{R}^d$. For example, if the agent is a manipulator in a production pipeline, then the manipulator with $3$ degrees of freedom can be represented as a point in a product space of $\mathbb{S}^1 \times \mathbb{S}^1 \times \mathbb{S}^1$. Each orientations of the arm can be represented as a $3$ tuple $(\beta_1, \beta_2, \beta_3)$.
However, describing agent motion with only position and orientation configurations are sometimes not enough. For example, a configuration, consisting of position and orientation of an aircraft, may not be in a collision condition with any of the surrounding environmental obstacles, but the magnitude and direction of velocity of the aircraft may be such that a collision is imminent. Therefore one should consider, for a generic problem scenario, not just the position and orientation of the agent, but the linear and angular velocities too.

This argument indicates that the decision theoretic planning has to be done in a state space, which is an augmentation of the configuration space along with linear and angular velocities. Topologically, this comprises of the configuration space along with its tangent bundle. Let us augment this definition of configuration space with its tangent bundle. A tangent bundle of $\mathbb{C}$ is defined as $T(\mathbb{C}) = \cup_{q \in \mathbb{C}}T_q(\mathbb{C})$, where $T_q(\mathbb{C})$ is the collection of all tangent vectors at $q$. 
%

The configuration space together with its tangent bundle is called state space $X$, in which a state $x \in X$ is simply defined as $x = (q, \dot{q})$. Holonomic constraints can be defined as $h_i(q,t) = 0$. Non-holonomic constraints require the use of rate variables and or inequalities, that is $l_i(q,\dot{q},t) = 0$ or $l_i(q,\dot{q},t) < 0$. Differential constraints can be written in Lagrangian dynamics as a set of equations of the form $g_i(q,\dot{q},\ddot{q},t) = 0$, additionally involving acceleration. These equations can be rewritten in the form $\dot{x} = f(x,u)$, where $u \in U$ the set of allowable control inputs to the system. The equations thus describe the state transitions resulting from a control input.

A configuration $q$ is free if the agent placed at $q$ does not collide with the obstacles, that is, $q$ does not belong to the set of points corresponding to the obstacles in the configuration space. Define the free space $\mathbb{C}_{\text{free}}$ to be the set of all free configurations in $\mathbb{C}$. We assume that there are a finite number of obstacles in the workspace or physical space, which are closed and bounded sets $\mathbb{O}_i$, $i=1,2,\ldots,m$, and without loss of generality, can be assumed to be pairwise disjoint. The configuration space obstacles can be defined as $\mathbb{C}_{\text{obs}} = \mathbb{C}\setminus \mathbb{C}_{\text{free}}$. However, we need to also define the state space obstacles.

To define state space obstacles apart from defining $x \in X_{\text{obs}} \Leftrightarrow q \in C_{\text{obs}}$ for $x = (q, (\dot q))$, we need to define the reachable set from an initial configuration. For the system defined by the expression $\dot{x} = f(x,u)$, a state $x'$ can be obtained by applying a control input $u$ over time $t$ from an initial state of $x_0$. The set of all possible $x'$ is called the reachable state of $x_0$ for a time $t$. For each state $x$, among the set of reachable states, one can define future collision states $X_{\text{fc}}$ and free state space $X_{\text{free}} = X \setminus X_{\text{fc}}$.
%
\subsection{Multi-Agents}
The extension of definitions from single agent to multi-agent is natural. Each agent is denoted as $A_i$, $i = 1,2,\ldots n$, where $n$ is the total number of agents. For simplicity we consider all agents are homogeneous. Each agent has a sensor scan radius of $r$. The dynamics of the agents can be defined as the following differential equation
\begin{equation}\label{System_Dynamics}
\dot{x}_{A_i} = f(x_{A_i},u_{A_i})    
\end{equation}
where $i \in \{1,2,\ldots n\}$, $x_{A_i} \in \mathbb{R}^{n_{A_i}}$ denoted the states of each agent and $u_{A_i} \in U \in \mathbb{R}^{m_{A_i}}$ denotes the control inputs. The state space of all the agents can be described as a product space of individual agents. That is ,
\begin{equation}
X = X_1 \times X_2 \ldots \times X_n
\end{equation}
Corresponding to each agent $A_i$, there exists a decision strategy space $D_i$. Each agent gathers information $y_i$ about the state of the environment $S$ through its sensors. Agents generates decision strategies according to the information obtained. That is,
\begin{equation}
D_i = G_i(y_i)
\end{equation}
where $G_i$ is the decision function.

For, $n$ agents or decision makers $A_1,A_2,\dots, A_n$, let a decision $a = (a_1,\ldots,a_n) \in D = D_1 \times \ldots \times D_n$, where $D$ is the set of all possible decisions, and each agent chooses a component decision $a_i$ from the possible decisions $D_i$. Let $\mathbb{D}_i$ be a $\sigma-$field of measurable subsets of $D_i$ for each $i = 1,\ldots,n$.
\subsection{The Environment}
The environment can be defined as $E \in \mathbb{R}^n$. The environment can be a physical closed space defined in $\mathbb{R}^2$/$\mathbb{R}^3$ or it can be a generic search domain defined in $\mathbb{R}^n, n > 3$. For example, in case of agents be as manipulators in a production pipeline with $10$ degree of freedom then the environment is defined as $E \in \mathbb{R}^{10}$ in which the manifold of manipulator configurations is embedded. In any case we assume there exists an appropriate transformation $T_E$ such that $E \in \mathbb{R}^n, n=2,3$ can be transformed into a $E_T \in \mathbb{R}^d$ where $d$ is the dimension of configuration space or state space. It can be said that $E$ or $E_T$ is $\textit{apriori}$ unknown and construction of $E_T$ is computationally expensive as this has exponential time and space complexity. We assume there exists a suitable metric $\rho$ for algebraic calculation in $E$ or in $E_T$.

Let, $\zeta \in \mathbb{R}^n$ denotes a random vector defined over an underlying probability space $(\mathbb{R}^n,\mathbb{F},P)$. For simplicity, we assume there exists an identification between the probability space and the environment $E_T$. The $\zeta$ represents the uncertainty of $E_T$ which are not controlled by any agents. However, the probability distribution of $\zeta$ is known to all the agents and it is assumed Gaussian. The uncertainty of the environment translates into distortion in information, on-the-fly discovery of non-reachable sets by the agents, disturbance such as wind to affect the system dynamics.
\subsection{The Pay-offs}
The pay-offs or reward can be defined as a real valued objective functional denoted as 
\begin{equation}
J: X \times D \rightarrow \mathbb{R}
\end{equation}
Here the environment states, the uncertainties and the states of the agents are reflected in $X$. $D$ is the joint decision space of all the agents as defined previously. Due to the probabilistic representations of the uncertainties in the environment $E_T$, the objective functional is essentially an expected payoff defined as
\begin{equation}
J = \mathbb{E}\{J(x_{A_1},a_{A_1},x_{A_2},a_{A_2},\ldots,x_{A_n},a_{A_n})\}
\end{equation}
Each specification of a decision $a$ transforms $J$ into a random variable defined on the probability space $\mathbb{R}^{\text{dim}(E_T)}$. The optimal decision maximizes(minimizes) the expectation of this random variable. The nature of the pay-off function may be or may not be available to all the agents. In a centralised problem the payoff is essentially defined over joint decision of the agents, where as in decentralised schemes the payoffs corresponds to individual agents decision strategy.
\subsection{Characteristics of Decision Strategy Space}
The optimal decision strategy is the one which minimizes expected pay-off $J$. That is,
\begin{equation}\label{final_equation}
 a^* = \argmin_{a \in D} J = \mathbb{E}\{J(x_{A_1},a_{A_1},x_{A_2},a_{A_2},\ldots,x_{A_n},a_{A_n})\}
\end{equation}
Here, $a^* \in D$. $D$ has defined structures, for example the agent decision function can be limited to discrete policies. These policies are basically control laws in the form of $u_{A_i} = K_{A_i}x_{A_i}$ which satisfies equation \ref{System_Dynamics}. Many often $J$ may not be a convex functional of $u_{A_i}, i=1,2,\ldots,n$ and hence a closed form solution may not be available. It is therefore necessary to discretise $U$ in order to choose an optimal strategy.

The co-operative multi-agent decision and control problem can therefore be defined as to find $u_{A_i}$ for each agent $A_i$, in a decentralised way which minimizes an expected pay-off defined as in equation \ref{final_equation}.

Next we present the integrated decision planning and control framework using RRT.
\section{An Integrated Decision and Control using RRT}\label{Integrated Planning}
As mentioned in Section \ref{Cooperative Search Framework}, an integrated co-operative planning of multi agent search must take into account of system dynamics while planning a decision. However, in most cases, the decision making or path planning  in a state space $X \in \mathbb{R}^d$ cannot be obtained as a closed form solution. Even without the consideration of the inner loop control and neglecting the system dynamics, the computational complexity of configuration space is exponential with the degrees of freedom of the agents. Therefore, an enumeration of possible decision strategies is computationally expensive and one should not be looking for such. This curse of dimensionality can be avoided by using randomized algorithms. However, according to No-Free-Launch-Theorem, randomized techniques comes with the price of algorithmic completeness and instead a probabilistic completeness can be achieved.

We used randomized algorithm such as Rapidly-Exploring-Random-Tree as a decision and control planner. For the sake of completeness we briefly outline the methodology of the algorithm.

Rapidly Exploring Random Tree (RRT) has been shown to be very effective in solving robot motion planning problems in a complex state space with kinodynamic motion constraints. RRT is introduced in \cite{lavalle}, \cite{lavalleandknuffer} as an efficient data structure and sampling scheme to quickly search high dimensional spaces that have algebraic constraints (arising from the obstacle) and differential constraints (arising from nonholonomy and system dynamics). The algorithm incrementally builds a tree whose nodes are different states of the robot$/$vehicle. These nodes are added randomly to the tree until one of the node comes close enough to any of the states in $x_{goal}$. Next, that goal state is added to the tree and a solution trajectory connecting $x_{goal}$ and $x_{init}$ can be found by backtracking the nodes. The edges of the tree forms a one feasible path or solution trajectory connecting a pair of initial and final states. The key idea behind RRT is to bias the tree growth towards unexplored regions of state space by random sampling and extending tree nodes to those regions. The selection of tree nodes for expansion is heavily dependent on current spatial distribution of tree nodes within the state space. Implicitly, the nodes with larger Voronoi cells\footnote{Voronoi diagram for a set of points $S$ in a plane is a partitioning of the plane with respect to those points. The partition is formed in such a way that each point in $S$ belongs to one partition. The points in $S$ are called sites or nodes. The distance between any point within a partition and the corresponding node is less than the distance between the point and any other node. For a definition of Voronoi Partition,} are more probable for extension. This is because the probability that a node is selected for expansion is directly proportional to the volume measure of its Voronoi cell. The tree node extension logic is based on forward simulation of system dynamics upon random control input.
In the following, we present the basic RRT algorithm. The RRT algorithm consists of two subroutines, Build-RRT \ref{Build RRT} and Extend-RRT \ref{Extend RRT}. Details of these algorithms can be found in \cite{choset}.

\begin{algorithm}
\SetAlgoLined
 $T\cdot{init}(x_{init})$~$\rhd$~Initialize tree $T$\;
 \For{i=1,\ldots,K}
 {
  $x_{rand}\leftarrow$ \text{Random Configuration}\;
  \text{Extend} $(T,x_{rand})$
 }
 \Return{$T$}
 \caption{Build-RRT}\label{Build RRT}
\end{algorithm}

\begin{algorithm}
\SetAlgoLined
 $x_{near}\leftarrow$\text{Nearest Neighbour} $(x,T)$\;
 $x_{new} \leftarrow$\text{New State}$(x_{near},u)$\;
 \If{$x_{new}$ \text{is Not in Obstacle}}
  {
   $T\cdot{add\_vertex} (x_{new})$\;
   $T\cdot{add\_edge} (x_{near},x_{new},u_{new})$\;
  \If{$x_{new} \in X_{goal}$}
  {
    \Return{Reached}\;    
    \Else
    {
    \Return{Continue}
    }
    }
    }
 \caption{Extend RRT}\label{Extend RRT}
\end{algorithm}

A random selection over possible decision strategies can reduce the cardinality of the strategy space $D$. An integrated decision planning and control demands that the integrated decision and control strategy space, hereafter mentioned as $D_I$ should consist of numerous feasible control strategies. One can assume that there exists a $f_I : D \rightarrow D_I$. The function $f_I$ is the control law for the two point boundary value problem. If the dynamics of the agents are linear and everywhere reachable then $f_I$ is an one-to-one mapping. Then to each $a_{A_i} \in D$ there exists a control law $u_{A_i}$ which maps $D$ to $D_I$. However, in the integrated decision planning and control framework, instead of finding $u_{A_i}$ for each $a_{A_i}$, the algorithm samples over $D_I$, that is $u_{A_i}$. In each iteration, for each agent, a set of feasible trajectories is generated using the RRT algorithm which samples different control input sequence from the feasible control input set. Without loss of generality we assume that there exists a binary collision checker which identifies collision states $X_{fc}$ and free state space $X_{free} = X \setminus X_{fc}$ for a arbitrary sequence of $u_{A_i}$.

\begin{figure}
  \centering
  {\includegraphics[height=4cm,width=4cm]{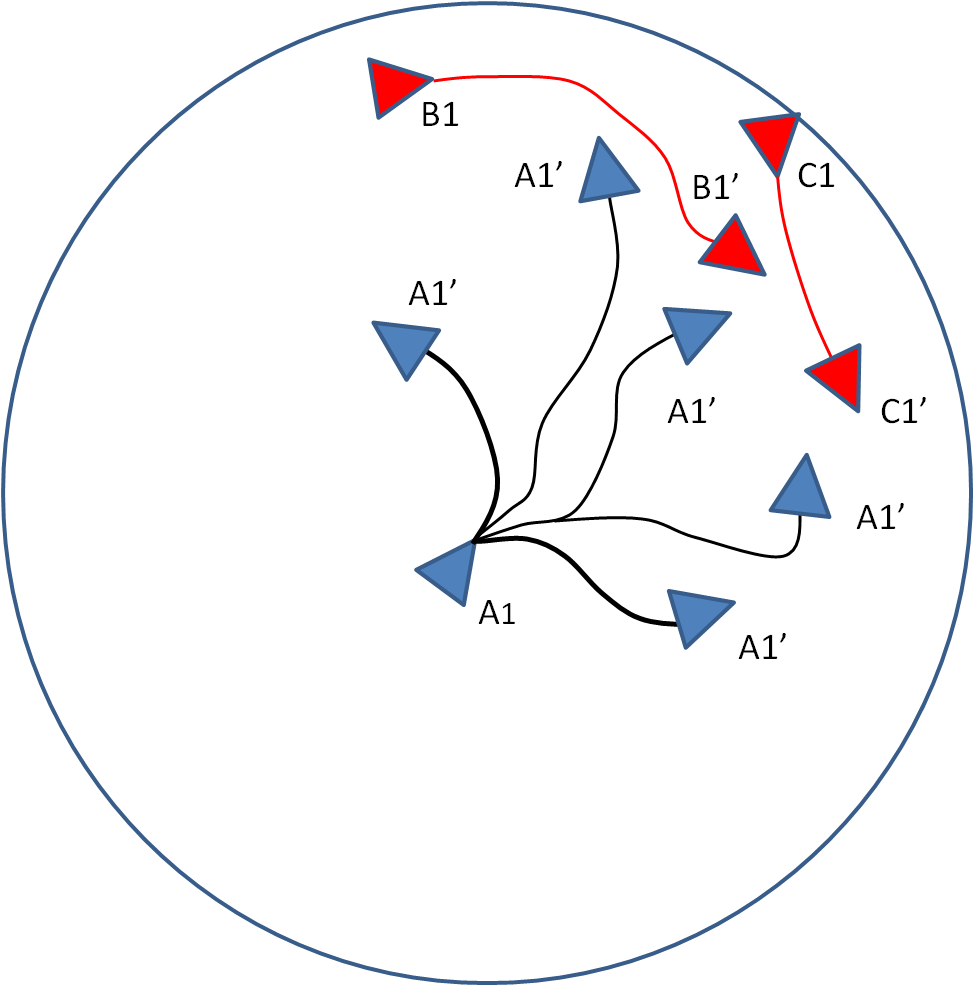}}
  \caption{Co-operative multi agent search: strategy combinations}
 \label{IntRRT}
\end{figure}

Note that, the agents do not communicate with each other and do not depend on others decision. This passive co-operation framework forces the $i^{\text{th}}$ agent to guess the possible trajectories of the agents within the sensor network. Then for the $i^{\text{th}}$ agent a set of feasible trajectories $S$ is generated. Let the number of trajectories or strategies is $S_q$. If there are $N_r$ number of agents exist within the sensor radius, then an obvious choice is to enumerate all these possible trajectories for each agent $A_j$, $j \in N_r$ and select the best trajectory by minimizing some performance index. The complexity of this technique is $\mathbb{O}((N_r+1)^{S_q})$. Such high complexity put a computational burden on each agent, especially when $N_r$ is high. To reduce the number of strategy enumeration, each agent guesses a best guessed trajectory for each of its neighbour and subsequently enumerates at most $\mathbb{O}(N_r S_q)$ number of strategies. 
For this, a flag value of $1$ is passed to the $\text{Build RRT}$ subroutine which extends only one branch of a RRT tree. However, for generating an agent's own strategies we use the RRT in its conventional mode, that is we pass a flag value of $0$ to the $\text{Build RRT}$ subroutine. Figure \ref{IntRRT} shows a pictorial view of strategy combinations and Algorithm \ref{IRRT} outlines the procedure for integrated decision and planning.

Note that, since the possible trajectories are created using RRT, therefore each single agent tries to search the entire search area. As RRT is Voronoi biased, the agents always pushed to the unexplored region of the state space. We assume that the agents is capable of storing the previous positional data and hence can update the uncertainty map of the environment. Therefore, an importance sampling can be used to sample random configurations from the state space as this will lead to faster exploration.

\begin{algorithm}
\SetAlgoLined
$R \leftarrow \text{Scan Radius}$\;
$N \leftarrow \text{Number of Agents}$\;
$d \leftarrow \text{No of Strategy}$\;
\For {i=1,\ldots,N}
{
  $M \rightarrow \text{Nearest.Neighbout}(i,R)$\;
  $U_i \rightarrow \text{Uncertainty Map of Agent i}$\;
  \For {j=1,\ldots,M}
  {
     $S_j \rightarrow \text{Guess Neighbour Strategy}(j,m_j|x_j^0,U_i)$\;
     $Q_j \rightarrow \text{Final.States}(S_j)$\;
  }
  \For {k=1,\ldots,d}
  {
     $T_k \rightarrow \text{Populate Self Strategy}(i,n_i|x_i^0)$\;
     $P_k \rightarrow \text{Final.States}(T_k)$\;
     $J_k \rightarrow \text{Frechet.Distance}(P_k,Q)$\;
  }
  $r_i \rightarrow \text{Select.Strategy}(\arg_k \max(J_k),T_k)$\;
  $x_i \rightarrow \text{Update.Agent State}(x_i^0,r_i)$\;
  $U_i \rightarrow \text{Update.UncertaintyMap}(U_i,x_i)$\;
}
\caption{Proposed Cooperative Search Algorithm}\label{IRRT}
\end{algorithm}

\begin{algorithm}
\SetAlgoLined
$l \rightarrow \text{No of RRT Nodes}$\;
$x_j^0 \rightarrow \text{Initial State of Agent j}$\;
$T_j \leftarrow x_j^0$\;
$U_i \rightarrow \text{Uncertainty Map of Agent i}$\;
\While {$k < l$}
{
	$T_j \leftarrow \text{Build-RRT}(T_j,U_i,X_{\text{goal}}= \varnothing,\text{Flag}=1)$\;

}
\caption{Guess Neighbour Strategy}\label{IRRT}
\end{algorithm}

\begin{algorithm}
\SetAlgoLined
$l \rightarrow \text{No of RRT Nodes}$\;
$x_i^0 \rightarrow \text{Initial State of Agent i}$\;
$T_i \rightarrow x_i^0$\;
$U_i \rightarrow \text{Uncertainty Map of Agent i}$\;
\While {$k < l$}
{
	$T_i \leftarrow \text{Build-RRT}(T_i,U_i,X_{\text{goal}}= \varnothing,\text{Flag}=0))$\;	
}
\caption{Populate Self Strategy}\label{IRRT}
\end{algorithm}
Various performance index can be used to compare the strategies among the agents. One can choose the strategy where there are minimum possible overlap with all the other agents. Since the agents do not communicate with each other, it is necessary to include intrinsically a similar cost metric when comparing the strategies and trajectories generated by those strategies. As an example, for $i^{\text{ith}}$ agent a cost functional can written as in terms of Frechet distance,
\begin{equation}
J_{ij, j\in N_r} = \inf_{\alpha, \beta} max_{t \in [0,1]}d\{p_{i\in S(i)}(\alpha(t)),p_{j\in S(N_r)}(\beta(t))\}
\end{equation}
where $\alpha(t)$ and $\beta(t)$ are two parametrization of the path $p_{i\in S_q}$ and $p_{j\in N_r}$, $t \in [0,1]$. As complexity of calculating the Frechet distance is $ \mathbb{O}(mn \cdot \log(mn))$ for two polygonal curves with $m$ and $n$ segments, a week Frechet distance can be calculated.

Another type of cost functional can be based on the coverage index of the area by the agents. The ideal coverage or search should prevent clustering of the agents in one area. Therefore, when comparing the trajectories the convex hull produced by $p_{ij}$, $i \in S(i), j\in S(N_r)$ can be evaluated. The strategy combination which yield the maximum convex hull area can be selected.
\section{Completeness of Cooperative Search Algorithm}\label{Analysis}
In \cite{yang2007multi}, a mathematical analysis of cooperative search method is proposed. Specifically it is shown that the method is probabilistically complete and an upper as well as lower bound is established on search time. It is shown that by using a cooperative strategy the agents essentially do not exhibit periodic paths. However, this is guaranteed only against a discretization of the search space. The method also relies on the representation of the uncertainties in the search space. We show in the following that our proposed method is also probabilistically complete. In this process we give the definition of $\tau$-probabilistic completeness show the completeness of the proposed method. This analysis is also independent of uncertainty representation and discretization of the search space.

We begin with the following assumptions.
\begin{enumerate}
\item All agents are homogeneous. All the agent has identical dynamics, that is $\dot{x^i} = f(x^i,u^i)$ for $i = 1,\ldots,N$.
\item The dynamics $\dot{x^i} = f(x^i,u^i)$ does not exhibit limit cycle. That is, there are no strange attractors within $X$.
\end{enumerate}
Let the state of the $i^{\text{th}}$ agent is defined as ${x_n}^i$ at the $n^{\text{th}}$ iteration of the algorithm. Let the reachable set defined at $n^{\text{th}}$ iteration of the $i^{\text{th}}$ agent is ${v_n}^{x_i} = \{x_f \in X|x_f = \int_{t_n}^{t_n + \delta}f(x_{n-1}^i,u^i)$. We can set up a relational class as $V = \{v_1 \rightarrow v_2 \rightarrow \ldots v_n\}$. Without loss of generalisation it can be assumed that $X \in \cup_{n \in \mathcal{Z}} v_n$. That is the entire state space is within the union of reachable states.

However, at each iteration we sample only a finite number of controls from $v_n^{x^i}$ as $u^i = \{u_1,u_2,\ldots,u_d\}$. Let's assume, for simplicity, that each time we are choosing from a finite pre-defined set of samples. 

Next we define the $\tau$-completeness of search algorithms. Let the outcome of a search algorithm is a parametrised curve passing through points in $X$. Let, $x' = \{x_n^i | n \in \mathcal{Z}\}$ set of points generated by a search algorithm after $n$ iterations. Let $\rho$ defines a distance metric defined on space $X$. With respect to the distance metric $\rho$, for any point $x_q \in X$, if $\inf\rho(x_q,x') <\tau$ as $n \rightarrow \infty$, then the search algorithm is $\tau$-probabilistically complete. Note  that, in the event $\tau \rightarrow 0$, any search algorithm is fundamentally incomplete.

Next we present the following theorem.
\begin{theorem}
The proposed search algorithm is $\tau$-probabilistically complete.
\end{theorem}
\begin{proof}
Let the neighbourhood radius is defined as $R$. We will show that completeness is ensured with the help of non-existence of any cyclic paths in $X$, when the agents cooperatively plans their paths. Specifically, we will show that if there are no limit cycles exists in the cooperative trajectory planning then as $n \rightarrow \infty$ the trajectories will be space filling.

Now as the individual agent does not show limit cycle behaviour then when the agent's decision is not affected by its neighbouring agents there will be no cyclic trajectories, which is trivial from the assumptions. However, depending on the magnitude of $R$ and the number of agents there can be two possibilities. 
\begin{itemize}
\item Some agents can be within the neighbourhood for a certain amount of time but not always. However, in this case when the agent is not influenced by any neighbour, the agent behaved according to its own dynamics and hence will show no limit cycle or cyclic segments in the generated trajectory.
\item The other possibility is that, some agents are always be present within the neighbourhood. This implies that individual agents path is always affected by at least one agent for all the time. Let us assume in this case an agent show indeed the existence of a limit cycle. This limit cycle corresponds to a input sequence $\{u_{k+1},u_{k+2},\dots u_{k+n}$ for some $k$ applied to an agent in order to obtained a cyclic trajectory upto the iteration $n$. Therefore any future periodic orbit requires application of these input in $k = T, 2T, \ldots$ time intervals.

Now $u_i$ can be obtained from any of the $d$ inputs. This implies that the probability of selecting $u_1$ at $k = 0$ is simply $(1/d)$. The probability of selecting $u_2$ at $k = 1$ given $u_1$ is selected at $k = 0$ is $(1/d^2)$. Hence for a $n$ iteration cyclic path probability of selecting the sequence $\{u_1,u_2,\ldots,u_n\}$ repeatedly is $(1/d^n)$. Let $E$ denotes the event that a sequence $\{u_1,u_2,\dots, u_n\}$ is chosen. Then
\begin{equation*}
\sum_{n = k+1}^{\infty}p(E_n) = \sum_{n= k+1}^{\infty}(1/d^n) < \infty
\end{equation*}
Since, $d > 1$.

Therefore, according to Borell-Cantelli Lemma $\lim_{n \rightarrow \infty}\sup P(E_n) = 0$. This implies that the event $E_n$ will not occur infinitely often. This implies there will be no periodic trajectory. Hence as $n \rightarrow \infty$, some point within the trajectory will eventually come as close to $x_q$ as $\tau$.
\end{itemize}
\end{proof}
\section{Results and Discussions}\label{Simulation}
\subsection{Simulation Setup}

\begin{figure}
  \centering
  {\includegraphics[scale = 0.2,keepaspectratio = true]{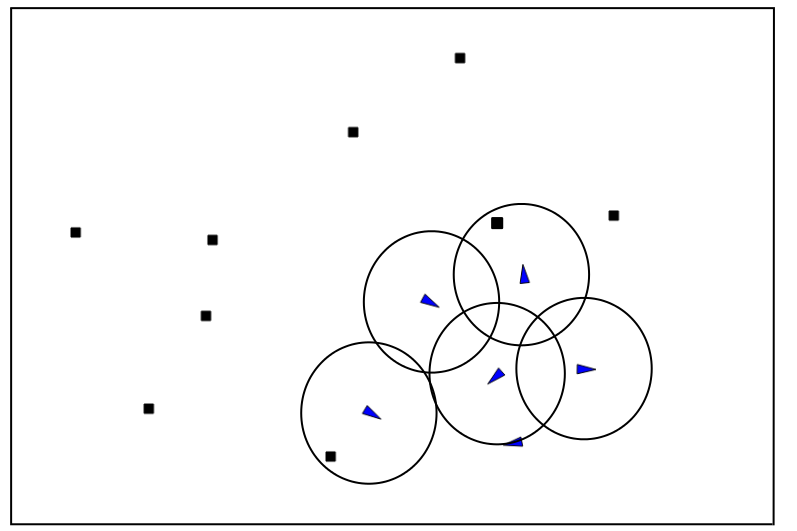}}
  \caption{An example scenario of co-operative multi-agent search consisting of agents ($n = 6$) and targets ($m = 8$)}
 \label{Example_Scenario}
\end{figure}

Simulations are performed for $n$ agents $(n = 6,8,10)$ to explore an unknown environment $\Omega$ containing $m$ targets $(m=8)$. One such arrangement is shown in Fig \ref{Example_Scenario}. The objective of the simulation is to show the feasibility in realization of the MAS-IDC approach. The performance of the MAS-IDC algorithm is also compared with Random Search methods.

The uncertainty of the environment is considered as dynamic and represented as a time functional. To analyse the uncertainty variations, the environment is decomposed into axis aligned rectangular grid cells, denoted as $\Xi_{ij}, i\in 20, j\in 10$.
The uncertainty of $\Xi_{ij}$ is defined as,
\begin{equation}
\eta_{\Xi_{ij}}(t) =  1 - e^{-(t-t')}
\end{equation}
where $t'$ is the time when $\Xi_{ij}$ last visited by any agent. The time interval between two successive visit is defined as the visiting period, $\tau_{\text{Vp}} = t - t'$. Since, $\eta_{\Xi_{ij}}$ is a time varying stochastic quantity, therefore a finite time average $M\left[\eta_{\Xi_{ij}}\right] $ and finite time variance $\text{Var}(\eta_{\Xi_{ij}})$ is used to characterise the uncertainty of each cell. That is,
\begin{eqnarray}
M\{\eta_{\Xi_{ij}}(t)\} = \frac{1}{T}\int_{0}^{T}\eta_{\Xi_{ij}}(t)dt \\
\text{Var}\{\eta_{\Xi_{ij}}(t)\}= \frac{1}{T}\int_{0}^{T}[\eta_{\Xi_{ij}}(t) - M\{\eta_{\Xi_{ij}}(t)\}]^{2}dt
\end{eqnarray}
$M'\left[\eta_{\Xi}\right]$, defined as the average of $M\left[\eta_{\Xi_{ij}}\right]$ over all the cells, is used to measure the performance of the the search algorithm. Although, we consider the position of the targets as static and uncertain, however the target strengths are time varying quantity, that is,
\begin{equation}
T_i(t) = 1 - e^{-(t-t_{T}')}
\end{equation}
The total target strength $T(t)$ is defined as the summation of the individual target strength.
\begin{equation}
T(t) = \sum_{i = 1}^{m} T_i(t)
\end{equation}
As agents we consider a simple non-holonomic dynamics represented by the following equations \ref{Car_Eq_1}-\ref{Car_Eq_3}. 
\begin{eqnarray}\label{Chapter4_Car_Equaitons}
\dot x &=& v~\cos \theta \label{Car_Eq_1} \\
\dot y &=& v~\sin \theta \label{Car_Eq_2}\\
\dot {\theta} &=& (v/L)~\tan\phi \label{Car_Eq_3}
\end{eqnarray}
where, $L$ is the length between the wheels, $v$ is the velocity, $\phi$ is the steering angle. The position of the agent and it's orientation is denoted by a $3$-tuple $(x,y,\theta)$. The control variables are velocity $v$ and steering angle. The configuration space is $\mathbb{R}^2\times S^1$. For example simulation such a simplistic vehicle equation is considered although any complex vehicle dynamics can be used in the algorithm.

\begin{figure}
  \centering
  {\includegraphics[scale = 1.0,keepaspectratio = true]{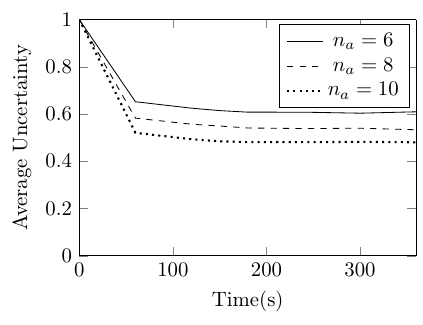}}
  \caption{Variation of Average Uncertainty $T(t)$ for $n=6$,$n=8$ and $n=10$ }
 \label{IntRRTUncertainty_1}
\end{figure}

\begin{figure}
  \centering
  {\includegraphics[scale = 1.0,keepaspectratio = true]{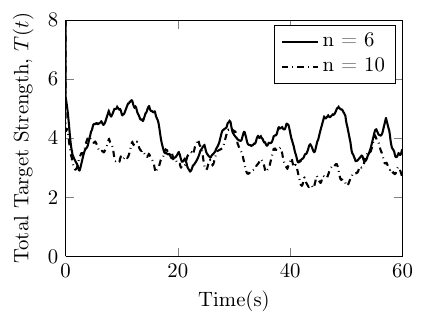}}
  \caption{Variation of Total Target Strength $T(t)$ for $n=6$ and $n=10$ }
 \label{IntRRT_1}
\end{figure}

\begin{figure}
  \centering
  {\includegraphics[scale = 1.0,keepaspectratio = true]{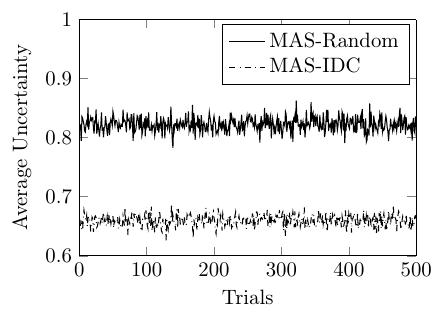}}
  \caption{Co-operative multi agent search: Average uncertainty comparison with Random search}
 \label{IntRRT_2}
\end{figure}

Various Monte-Carlo simulations are performed considering various parameter variations such as sensing radius, number of agents etc. In each case a set of $500$ simulation trials are performed and results are obtained. One such example scenario is shown in Fig. \ref{Example_Scenario}. The sensor radius considered here is $\unit[20]{m}$. The environment considered here as a rectangular region with $\unit[100]{m} \times \unit[150]{m}$ dimensions. The variation of average uncertainty of the area is shown in Fig.\ref{IntRRTUncertainty_1} and the variation of total target strength is shown in Fig. \ref{IntRRT_1} for $n=6$ and $n = 10$ agents. Fig. \ref{IntRRT_2} shows considerable improvement when comparing with random search methodology. Note that, we only show a comparison of our method with a basic random search method suitably modified to include the dynamics of the vehicle. This is needed as other co-operative decentralised search without communication do not include vehicle dynamics into account. 

\begin{figure}
  \centering
  {\includegraphics[scale = 1.0,keepaspectratio = true]{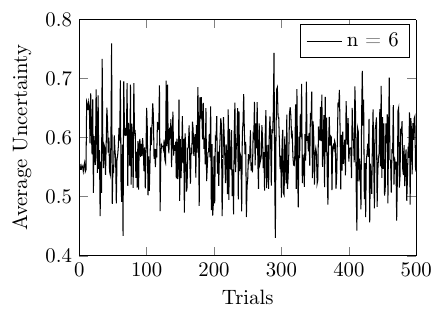}}
    \caption{Average Cell uncertainty when $\sigma_{\text{Error}} = 0$}
 \label{Sigma_0}
\end{figure}

\begin{figure}
  \centering
  {\includegraphics[scale = 1.0,keepaspectratio = true]{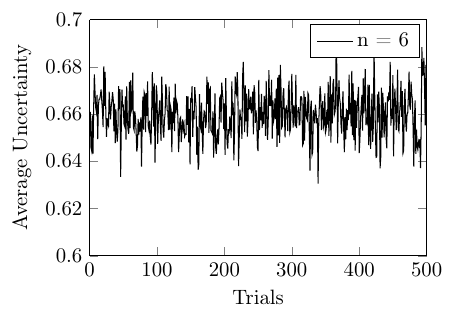}}
    \caption{Average Cell uncertainty when $\sigma_{\text{Error}} = 2$}
 \label{Sigma_1}
\end{figure}

\begin{figure}
  \centering
  {\includegraphics[scale = 1.0,keepaspectratio = true]{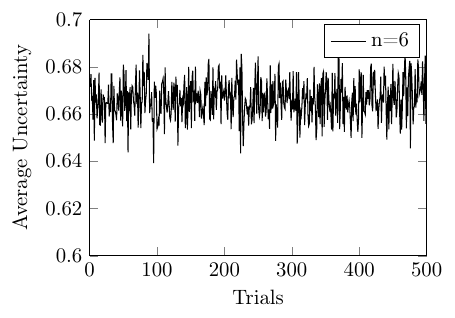}}
    \caption{Average Cell uncertainty when $\sigma_{\text{Error}} = 4$}
 \label{Sigma_2}
\end{figure}

We assume that each agent has sensors to measure the position and velocity of the other agents. However, there could be error in estimation of such states. We consider a $\pm \unit[2]{m}$ to $\pm \unit[4]{m}$ positional error variance in the estimation. The resulting average cell uncertainties are shown in Fig.\ref{Sigma_0}, Fig.\ref{Sigma_1}, and Fig.\ref{Sigma_2}. These show that the average uncertainties do not changes drastically with the inaccuracy in the estimation to some degree.



\begin{figure}
  \centering
  {\includegraphics[scale = 1.0,keepaspectratio = true]{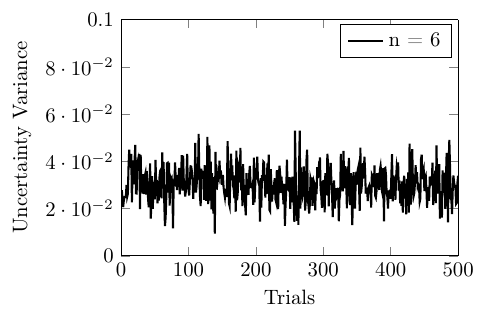}}
  \caption{Cell uncertainty variance when $\sigma_{\text{Error}} = 0$}
 \label{Var_1}
\end{figure}

\begin{figure}
  \centering
  {\includegraphics[scale = 1.0,keepaspectratio = true]{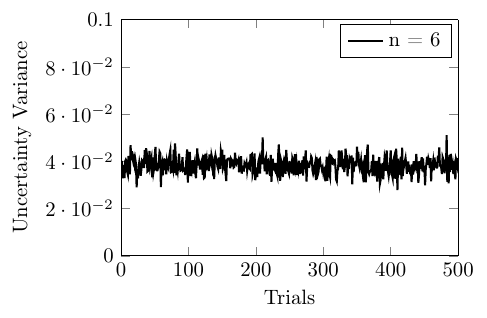}}
  \caption{Cell uncertainty variance when $\sigma_{\text{Error}} = 2$}
 \label{Var_2}
\end{figure}

\begin{figure}
  \centering
  {\includegraphics[scale = 1.0,keepaspectratio = true]{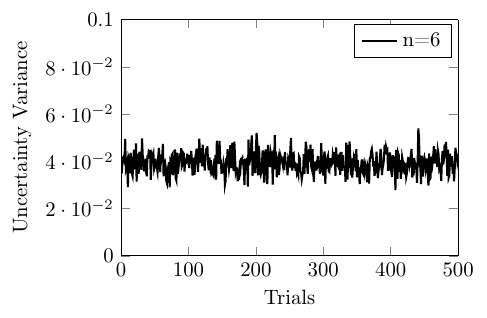}}
  \caption{Cell uncertainty variance when $\sigma_{\text{Error}} = 4$}
 \label{Var_3}
\end{figure}

We also show the resulting variance in the environmental uncertainties as in Fig.\ref{Var_1}, Fig.\ref{Var_2}, Fig. \ref{Var_3}. The relatively small values implies that the environment is been searched in an uniform manner despite the positional inaccuracy. We also see that despite of knowing the accurate intentions of the neighbourhood agents, the number of collision between the agents in case of IDC approach, is quite small, on an average of $0.7$ over the trials, while for the case of Random search, it is $6$.

\section{Conclusion}\label{Conclusion}
In this paper we show a framework for integrated decision planning and control for multi-agent search problem using randomized algorithms, specifically RRT. The method is iterative and applicable to a wide class of problems and complex systems. We show that the algorithm is $\tau$ probabilistically complete. We also present simulations results which show considerable improvement over the random search. 

\bibliographystyle{IEEEtran}

\bibliography{Reference}
\end{document}